# Point Set Registration: Coherent Point Drift

Andriy Myronenko and Xubo Song

**Abstract**—Point set registration is a key component in many computer vision tasks. The goal of point set registration is to assign correspondences between two sets of points and to recover the transformation that maps one point set to the other. Multiple factors, including an unknown non-rigid spatial transformation, large dimensionality of point set, noise and outliers, make the point set registration a challenging problem. We introduce a probabilistic method, called the Coherent Point Drift (CPD) algorithm, for both rigid and non-rigid point set registration. We consider the alignment of two point sets as a probability density estimation problem. We fit the GMM centroids (representing the first point set) to the data (the second point set) by maximizing the likelihood. We force the GMM centroids to move *coherently* as a group to preserve the topological structure of the point sets. In the rigid case, we impose the coherence constraint by re-parametrization of GMM centroid locations with rigid parameters and derive a closed form solution of the maximization step of the EM algorithm in arbitrary dimensions. In the non-rigid case, we impose the coherence constraint by regularizing the displacement field and using the variational calculus to derive the optimal transformation. We also introduce a fast algorithm that reduces the method computation complexity to linear. We test the CPD algorithm for both rigid and non-rigid transformations in the presence of noise, outliers and missing points, where CPD shows accurate results and outperforms current state-of-the-art methods.

**Index Terms**—Registration, correspondence, matching, alignment, rigid, non-rigid, point sets, Coherent Point Drift (CPD), Gaussian mixture model (GMM), coherence, regularization, EM algorithm.

## 1 INTRODUCTION

REGISTRATION of point sets is a key component in many computer vision tasks including stereo matching, content-based image retrieval, image registration and shape recognition. The goal of point set registration is to assign correspondences between two sets of points and/or to recover the transformation that maps one point set to the other. For example, in stereo matching, in order to recover depth and infer structure from a pair of stereo images, it is necessary to first define a set of points in each image and find the correspondence between them. An example of point set registration problem is shown in Fig. 1. The "points" in a point set are often features extracted from an image, such as the locations of corners, boundary points or salient regions. The points can represent both geometric and intensity characteristics of an image.

Practical point set registration algorithms should have several desirable properties: (1) Ability to accurately model the transformation required to align the point sets with tractable computational complexity; (2) Ability to handle possibly high dimensionality of the point sets; (3) Robustness to degradations such as noise, outliers and missing points that occur due to imperfect image acquisition and feature extraction.

The transformation usually falls into two categories: rigid or non-rigid. A rigid transformation allows only for translation, rotation and scaling. The simplest non-rigid transformation is affine, which also allows anisotropic scaling and skews. Non-rigid transformation occurs in many real-world problems including deformable motion tracking, shape recognition and medical image registration. The true underlying non-rigid transformation model is often unknown and challenging to model. Simplistic approximations of the true non-rigid transformation, including piece-wise affine and polynomial models, are often inadequate for correct alignment and can produce erroneous correspondences. Due to the usually large number of transformation parameters, the non-rigid point sets registration methods tend to be sensitive to noise and outliers and are likely to converge into local minima. They also tend to have a high computational complexity. A practical non-rigid point set registration method should be able to accurately model the non-rigid transformation with tractable computational complexity.

Multidimensional point sets are common in many real world problems. Most current rigid and non-rigid point sets registration algorithm are well suited for 2D and 3D cases, but their generalization to higher dimensions are not always trivial. Furthermore, degradations such as noise, outliers and missing points significantly complicates the problem. Outliers are the points that are incorrectly extracted from the image; outliers have no correspondences in the other point set. Missing points are the features that are not found in the image due to occlusion or inaccurate feature extraction. A point set registration method should be robust to these degradations.

We present a robust probabilistic multidimensional point sets registration algorithm for both rigid and non-rigid transforms. We consider the alignment of two point sets as a probability density estimation problem,

- A. Myronenko and X. Song are with the Department of Science and Engineering, School of Medicine, Oregon Health and Science University, Portland, OR, 97201.
  E-mail: myron@csee.ogi.edu, xubosong@csee.ogi.edu

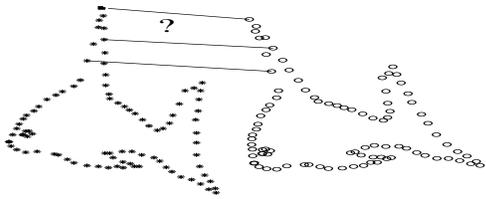

Fig. 1. The point set registration problem: Given two sets of points, assign the correspondences and the transformation that maps one point set to the other.

where one point set represents the Gaussian Mixture Model (GMM) centroids, and the other one represents the data points. We fit the GMM centroids to the data by maximizing the likelihood. At the optimum, the point sets become aligned and the correspondence is obtained using the posterior probabilities of the GMM components. Core to our method is to force GMM centroids to move *coherently* as a group, which preserves the topological structure of the point sets. We impose the coherence constraint by explicit re-parametrization of GMM centroid locations (for rigid and affine transformations) or by regularization of the displacement field (for smooth non-rigid transformation). We also show how the computational complexity of the method can be reduced to linear, which makes it applicable for large data sets. The rest of the paper is organized as follows. In Section 2, we overview the current rigid and non-rigid point set registration methods and state our contributions. In Section 3, we formulate a probabilistic point set registration framework. In Sections 4 and 5, we describe our algorithms for rigid, affine and non-rigid registration cases, and relate them to existing works. In Section 6, we discuss the computational complexity of the method and introduce its fast implementation. In Section 7, we evaluate the performance of our algorithm. Section 8 concludes with some discussions.

## 2 Previous Work

Many algorithms exist for rigid and for non-rigid point set registration. They aim to recover the correspondence or the transformation required to align the point sets or both. Many algorithms involve a dual-step update, iteratively alternating between the correspondence and the transformation estimation. Here, we briefly overview the rigid and non-rigid point set registration methods and state our contributions.

### 2.1 Rigid Point set Registration Methods

Iterative Closest Point (ICP) algorithm, introduced by Besl and McKay [1] and Zhang [2], is the most popular method for rigid point set registration due to its simplicity and low computational complexity. ICP iteratively assigns correspondences based on a closest distance criterion and finds the least-squares rigid transformation relating the two point sets. The algorithm then redetermines the correspondences and continues until it reaches the local minimum. Many variants of ICP have been proposed that affect all phases of the algorithm from the selection and matching of points to the minimization strategy [3], [4]. ICP requires that the initial position of the two point sets be adequately close.

To overcome the ICP limitations, many probabilistic methods were developed [5], [6]. These methods use soft-assignment of correspondences that establishes correspondences between all combinations of points according to some probability, which generalizes the binary assignment of correspondences in ICP. Among these methods are Robust Point Matching (RPM) algorithm introduced by Gold et al. [7], and its later variants [5], [8], [9]. In [10] it was shown that in RPM alternating soft-assignment of correspondences and transformation is equivalent to the Expectation Maximization (EM) algorithm for GMM, where one point sets is treated as GMM centroids with equal isotropic covariances and the other point set is treated as data points. In fact, several rigid point set methods, including Joshi and Lee [11], Wells [12], Cross and Hancock [13], Luo and Hancock [6], [14], McNeill and Vijayakumar [15] and Sofka et al. [16], explicitly formulate the point sets registration as a maximum likelihood (ML) estimation problem, to fit the GMM centroids to the data points. These methods re-parameterize GMM centroids by a set of rigid transformation parameters (translation and rotation). The EM algorithm used for optimization of the likelihood function consists of two steps: E-step to compute the probabilities, M-step to update the transformation. Common to such probabilistic methods is the inclusion of an extra distribution term to account for outliers (e.g. large Gaussian [5] or uniform distribution [12]) and deterministic annealing on the Gaussian width to avoid poor local minima. These probabilistic methods perform better than conventional ICP, especially in presence of noise and outliers.

Another class of rigid point sets registration methods are the spectral methods. Scott and Longuet-Higgins [17] introduced a non-iterative algorithm to associate points of two arbitrary patterns, exploiting some properties of Gaussian proximity matrix (Gram matrix) of point sets. The algorithm works well with translation, shearing and scaling deformations, but performs poorly with non-rigid transformations. Li and Hartley showed that correspondence and transformation are two factors of Gram matrices, and can be found iteratively using Newton-Schulz factorization [18]. This method performs well for moderate linear transformations. In spite of its elegance, the large computational effort required for spectral methods prohibits its wide applicability. There are a few other nonspectral methods worth mentioning. Ho et al. [19] proposed an elegant non-iterative algorithm for 2D affine registration by searching for the roots of the associated polynomials. Unfortunately this method does not generalize to higher dimensions. Belongie et al. [20] introduced the "shape context" descriptor, which incorporates the neighborhood structure of the point set and thus helps to recover the correspondence between the point sets.

Our approach to the rigid point sets registration is probabilistic and most closely related to the works of Rangarajan et al. [5], Wells [12] and Luo and Hancock [14]. Despite extensive work in rigid probabilistic registration, none of the methods, to our best knowledge, provides a closed form solution to the maximization step (M-step) of the EM algorithm for a general multidimensional case. The fact that the rotation matrix should be constrained to be orthogonal and to have a positive determinant further complicates its estimation. Rangarajan and collaborators [5] showed the solution for 2D case only, where rotation is parametrized by a single angle. In higher dimensions the closed form solution with Euler angles parametrization is not feasible. Luo and Hancock [6], [14] find the rotation matrix through singular value decomposition (SVD). They ignored some terms of the objective function, which leads to only an approximate solution. We shall derive the exact closed form solution (M-step) for the rigid point set registration and discuss its difference from the related methods in Section 4.

### 2.2 Non-rigid Point set Registration Methods

Earlier works on non-rigid point set registration include Hinton et al. [21], [22], who used the probabilistic GMM formulation. The GMM centroids are uniformly positioned along the contour (modeled using splines), which allows for non-rigid transformations. In practice, the method is applicable only to contour-like point sets. One of the most popular non-rigid point set registration method is by Chui and Rangarajan [8], [9]. They proposed to use Thin Plate Spline (TPS) [23], [24] parametrization of the transformation, following RPM, which results into the TPS-RPM method. Similar to the rigid case, they use deterministic annealing and alternate updates for soft-assignment and TPS parameters estimation. They also showed that TPS-RPM is equivalent (with several modifications) to EM for GMM [10]. Tsin and Kanade [25] proposed a correlation-based approach to point set registration, which was later improved by Jian and Vemuri [26]. The method considers the registration as an alignment between two distributions, where each of the point sets represents the GMM centroids. One of the point sets is parametrized by rigid/affine parameters (in rigid/affine case) or TPS (in non-rigid case). The transformation parameters are estimated to minimize the $L_2$ norm between the distributions. These methods all use explicit TPS parametrization, which is equivalent to a regularization of second order derivatives of the transformation [23], [24]. The TPS parametrization does not exist when the dimension of points is higher than three, which limits the applicability of such methods.

Huang et al. [27] proposed to implicitly embed the shape (or point sets in our case) into distance transform space, and align them similar to non-rigid image registration algorithms. The authors use sum-of-squared-differences similarity measure between the embedded spaces and incremental free form deformation (FFD) to parameterize the transformation. The method performs well on relatively simple data sets.

Finally, in our previous work we presented the Coherent Point Drift (CPD) algorithm [28] for non-rigid point sets registration. The algorithm regularizes the displacement (velocity) field between the point sets following the motion coherence theory (MCT) [29], [30]. Using variational calculus, we obtained that the optimal displacement field has an elegant kernel form in multiple dimenstions. In this paper, we shall elaborate and analyze the CPD algorithm. We also extend the general non-rigid registration framework, and show that CPD and TPS-RPM are its special cases. Among other contributions, we estimate the width of GMM components without the time consuming deterministic annealing and show a fast CPD implementation to reduce the computational complexity to linear. We shall discuss and compare our method to the works of Chui and Rangarajan [9], and Jian and Vemuri [26] in Section 5.

## 3 GENERAL METHODOLOGY

We consider the alignment of two point sets as a probability density estimation problem, where one point set represents the Gaussian mixture model (GMM) centroids, and the other one represents the data points. At the optimum, two point sets become aligned and the correspondence is obtained using the maximum of the GMM posterior probability for a given data point. Core to our method is to force GMM centroids to move coherently as a group to preserve the topological structure of the point sets. Throughout the paper we use the following notations:

- $D$ - dimension of the point sets,
- $N, M$ - number of points in the point sets,
- $\mathbf{X}_{N \times D} = (\mathbf{x}_1, \ldots, \mathbf{x}_N)^T$ - the first point set (the data points),
- $\mathbf{Y}_{M \times D} = (\mathbf{y}_1, \ldots, \mathbf{y}_M)^T$ - the second point set (the GMM centroids),
- $\mathcal{T}(\mathbf{Y}, \theta)$ - Transformation $\mathcal{T}$ applied to $\mathbf{Y}$, where $\theta$ is a set of the transformation parameters,
- $\mathbf{I}$ - identity matrix,
- $\mathbf{1}$ - column vector of all ones,
- $\mathrm{d}(\mathbf{a})$ - diagonal matrix formed from the vector $\mathbf{a}$.

We consider the points in $\mathbf{Y}$ as the GMM centroids, and the points in $\mathbf{X}$ as the data points generated by the GMM. The GMM probability density function is

$$p(\mathbf{x}) = \sum_{m=1}^{M+1} P(m)p(\mathbf{x}|m) \qquad (1)$$

where $p(\mathbf{x}|m) = \frac{1}{(2\pi\sigma^2)^{D/2}} \exp^{-\frac{\|\mathbf{x}-\mathbf{y}_m\|^2}{2\sigma^2}}$. We also added an additional uniform distribution $p(\mathbf{x}|M+1) = \frac{1}{N}$ to the mixture model to account for noise and outliers. We use equal isotropic covariances $\sigma^2$ and equal membership probabilities $P(m) = \frac{1}{M}$ for all GMM components ($m = 1, \ldots, M$). Denoting the weight of the uniform distribution as $w$, $0 \leq w \leq 1$, the mixture model takes the form

$$p(\mathbf{x}) = w\frac{1}{N} + (1-w)\sum_{m=1}^{M} \frac{1}{M} p(\mathbf{x}|m) \quad (2)$$

We re-parameterize the GMM centroid locations by a set of parameters $\theta$ and estimate them by maximizing the likelihood, or equivalently by minimizing the negative log-likelihood function

$$E(\theta, \sigma^2) = -\sum_{n=1}^{N} \log \sum_{m=1}^{M+1} P(m)p(\mathbf{x}|m) \quad (3)$$

where we make the i.i.d. data assumption. We define the correspondence probability between two points $\mathbf{y}_m$ and $\mathbf{x}_n$ as the posterior probability of the GMM centroid given the data point: $P(m|\mathbf{x}_n) = P(m)p(\mathbf{x}_n|m)/p(\mathbf{x}_n)$.

We use Expectation Maximization (EM) algorithm [31], [32] to find $\theta$ and $\sigma^2$. The idea of EM is first to guess the values of parameters ("old" parameter values) and then use the Bayes' theorem to compute *a posteriori* probability distributions $P^{old}(m|\mathbf{x}_n)$ of mixture components, which is the expectation or E-step of the algorithm. The "new" parameter values are then found by minimizing the expectation of the complete negative log-likelihood function [32]

$$Q = -\sum_{n=1}^{N} \sum_{m=1}^{M+1} P^{old}(m|\mathbf{x}_n) \log(P^{new}(m)p^{new}(\mathbf{x}_n|m)) \quad (4)$$

with respect to the "new" parameters, which is called the maximization or M-step of the algorithm. The $Q$ function, which we call *the objective function*, is also an upper bound of the negative log-likelihood function in (3). The EM algorithm proceeds by alternating between E- and M-steps until convergence. Ignoring the constants independent of $\theta$ and $\sigma^2$, we rewrite (4) as

$$Q(\theta, \sigma^2) = \frac{1}{2\sigma^2} \sum_{n=1}^{N} \sum_{m=1}^{M} P^{old}(m|\mathbf{x}_n) \|\mathbf{x}_n - \mathcal{T}(\mathbf{y}_m, \theta)\|^2 + \frac{N_\mathbf{P} D}{2} \log \sigma^2 \quad (5)$$

where $N_\mathbf{P} = \sum_{n=1}^{N} \sum_{m=1}^{M} P^{old}(m|\mathbf{x}_n) \leq N$ (with $N = N_\mathbf{P}$ only if $w = 0$) and $P^{old}$ denotes the posterior probabilities of GMM components calculated using the previous parameter values:

$$P^{old}(m|\mathbf{x}_n) = \frac{\exp^{-\frac{1}{2}\left\|\frac{\mathbf{x}_n - \mathcal{T}(\mathbf{y}_m, \theta^{old})}{\sigma^{old}}\right\|^2}}{\sum_{k=1}^{M} \exp^{-\frac{1}{2}\left\|\frac{\mathbf{x}_n - \mathcal{T}(\mathbf{y}_k, \theta^{old})}{\sigma^{old}}\right\|^2} + c} \quad (6)$$

where $c = (2\pi\sigma^2)^{D/2} \frac{w}{1-w} \frac{M}{N}$. Minimizing the function $Q$, we necessarily decrease the negative log-likelihood function $E$, unless it is already at a local minimum. To proceed, we specify the transformation $\mathcal{T}$ for rigid, affine and non-rigid point set registration cases separately.

## 4 RIGID & AFFINE POINT SET REGISTRATION

For rigid point set registration, we define the transformation of the GMM centroid locations as $\mathcal{T}(\mathbf{y}_m; \mathbf{R}, \mathbf{t}, s) = s\mathbf{R}\mathbf{y}_m + \mathbf{t}$, where $\mathbf{R}_{D \times D}$ is a rotation matrix, $\mathbf{t}_{D \times 1}$ is a translation vector and $s$ is a scaling parameter. The objective function (5) takes the form:

$$Q(\mathbf{R}, \mathbf{t}, s, \sigma^2) = \frac{1}{2\sigma^2} \sum_{m,n=1}^{M,N} P^{old}(m|\mathbf{x}_n) \|\mathbf{x}_n - s\mathbf{R}\mathbf{y}_m - \mathbf{t}\|^2$$
$$+ \frac{N_\mathbf{P} D}{2} \log \sigma^2, \quad \text{s.t. } \mathbf{R}^T\mathbf{R} = \mathbf{I}, \ \det(\mathbf{R}) = 1 \quad (7)$$

Note that the first term is similar to the one in the absolute orientation problem [33], [34], which is defined as $min \sum_{n=1}^{N} \|\mathbf{x}_n - (s\mathbf{R}\mathbf{y}_n + \mathbf{t})\|^2$ in our notations. Equation (7) can be seen as a generalized weighted absolute orientation problem, because it includes weighted differences between all combinations of points. The exact minimization solution of the objective function (7) is complicated due to the constraints on $\mathbf{R}$. To obtain the closed form solution we shall use Lemma 1 [35]:

*Lemma 1:* Let $\mathbf{R}_{D \times D}$ be an unknown rotation matrix and $\mathbf{A}_{D \times D}$ be a known real square matrix. Let $\mathbf{U}SS\mathbf{V}^T$ be a Singular Value Decomposition (SVD) of $\mathbf{A}$, where $\mathbf{U}\mathbf{U}^T = \mathbf{V}\mathbf{V}^T = \mathbf{I}$ and $SS = \text{d}(s_i)$ with $s_1 \geq s_2 \geq, \ldots, \geq s_D \geq 0$. Then the optimal rotation matrix $\mathbf{R}$ that maximizes $\text{tr}(\mathbf{A}^T\mathbf{R})$ is $\mathbf{R} = \mathbf{U}\mathbf{C}\mathbf{V}^T$, where $\mathbf{C} = \text{d}(1, 1, \ldots, 1, \det(\mathbf{U}\mathbf{V}^T))$.

To apply this lemma, we need to simplify the $Q$ function to a form equivalent to $\text{tr}(\mathbf{A}^T\mathbf{R})$. First, we eliminate translation $\mathbf{t}$ from $Q$. Taking partial derivative of $Q$ with respect to $\mathbf{t}$ and equate it to zero, we obtain:

$$\mathbf{t} = \frac{1}{N_\mathbf{P}} \mathbf{X}^T \mathbf{P}^T \mathbf{1} - s\mathbf{R} \frac{1}{N_\mathbf{P}} \mathbf{Y}^T \mathbf{P} \mathbf{1} = \mu_\mathbf{x} - s\mathbf{R}\mu_\mathbf{y},$$

where the matrix $\mathbf{P}$ has elements $p_{mn} = P^{old}(m|\mathbf{x}_n)$ in (6) and the mean vectors $\mu_\mathbf{x}$ and $\mu_\mathbf{y}$ are defined as:

$$\mu_\mathbf{x} = \mathbf{E}(\mathbf{X}) = \frac{1}{N} \mathbf{X}^T \mathbf{P}^T \mathbf{1}, \quad \mu_\mathbf{y} = \mathbf{E}(\mathbf{Y}) = \frac{1}{N} \mathbf{Y}^T \mathbf{P} \mathbf{1}.$$

Substituting $\mathbf{t}$ back into the objective function and rewritting it in matrix form, we obtain

$$Q = \frac{1}{2\sigma^2}[\text{tr}(\hat{\mathbf{X}}^T \text{d}(\mathbf{P}^T\mathbf{1})\hat{\mathbf{X}}) - 2s\,\text{tr}(\hat{\mathbf{X}}^T \mathbf{P}^T \hat{\mathbf{Y}} \mathbf{R}^T) +$$
$$s^2 \text{tr}(\hat{\mathbf{Y}}^T \text{d}(\mathbf{P}\mathbf{1})\hat{\mathbf{Y}})] + \frac{N_\mathbf{P} D}{2} \log \sigma^2 \quad (8)$$

where $\hat{\mathbf{X}} = \mathbf{X} - \mathbf{1}\mu_\mathbf{x}^T$ and $\hat{\mathbf{Y}} = \mathbf{Y} - \mathbf{1}\mu_\mathbf{y}^T$ are the centered point set matrices. We use the fact that trace is invariant under cyclic matrix permutations and $\mathbf{R}$ is orthogonal. We can rewrite (8) as $Q = -c_1 \text{tr}((\hat{\mathbf{X}}^T \mathbf{P}^T \hat{\mathbf{Y}})^T \mathbf{R}) + c_2$, where $c_1, c_2$ are constants independent of $\mathbf{R}$ and $c_1 > 0$. Thus minimization of $Q$ with respect to $\mathbf{R}$ is equivalent to maximization of

$$\max \text{tr}(\mathbf{A}^T \mathbf{R}), \mathbf{A} = \hat{\mathbf{X}}^T \mathbf{P}^T \hat{\mathbf{Y}}, \quad \text{s.t. } \mathbf{R}^T\mathbf{R} = \mathbf{I}, \ \det(\mathbf{R}) = 1.$$

**Rigid point set registration algorithm:**
- Initialization: $\mathbf{R} = \mathbf{I}, \mathbf{t} = 0, s = 1, 0 \leq w \leq 1$
  $\sigma^2 = \frac{1}{DNM} \sum_{n=1}^{N} \sum_{m=1}^{M} \|\mathbf{x}_n - \mathbf{y}_m\|^2$
- EM optimization, repeat until convergence:
  - E-step: Compute $\mathbf{P}$,
  $$p_{mn} = \frac{\exp^{-\frac{1}{2\sigma^2}\|\mathbf{x}_n - (s\mathbf{R}\mathbf{y}_m + \mathbf{t})\|^2}}{\sum_{k=1}^{M} \exp^{-\frac{1}{2\sigma^2}\|\mathbf{x}_n - (s\mathbf{R}\mathbf{y}_k + \mathbf{t})\|^2} + (2\pi\sigma^2)^{D/2} \frac{w}{1-w} \frac{M}{N}}$$
  - M-step: Solve for $\mathbf{R}, s, \mathbf{t}, \sigma^2$:
    - $N_\mathbf{P} = \mathbf{1}^T \mathbf{P} \mathbf{1}, \mu_\mathbf{x} = \frac{1}{N_\mathbf{P}} \mathbf{X}^T \mathbf{P} \mathbf{1}, \mu_\mathbf{y} = \frac{1}{N_\mathbf{P}} \mathbf{Y}^T \mathbf{P} \mathbf{1}$,
    - $\hat{\mathbf{X}} = \mathbf{X} - \mathbf{1}\mu_\mathbf{x}^T, \hat{\mathbf{Y}} = \mathbf{Y} - \mathbf{1}\mu_\mathbf{y}^T$,
    - $\mathbf{A} = \hat{\mathbf{X}}^T \mathbf{P}^T \hat{\mathbf{Y}}$, compute SVD of $\mathbf{A} = \mathbf{U}SS\mathbf{V}^T$,
    - $\mathbf{R} = \mathbf{U}\mathbf{C}\mathbf{V}^T$, where $\mathbf{C} = d(1,..,1, \det(\mathbf{U}\mathbf{V}^T))$,
    - $s = \frac{\mathrm{tr}(\mathbf{A}^T \mathbf{R})}{\mathrm{tr}(\hat{\mathbf{Y}}^T d(\mathbf{P}\mathbf{1})\hat{\mathbf{Y}})}$,
    - $\mathbf{t} = \mu_\mathbf{x} - s\mathbf{R}\mu_\mathbf{y}$,
    - $\sigma^2 = \frac{1}{N_\mathbf{P} D} (\mathrm{tr}(\hat{\mathbf{X}}^T d(\mathbf{P}^T \mathbf{1})\hat{\mathbf{X}}) - s\,\mathrm{tr}(\mathbf{A}^T \mathbf{R}))$.
- The aligned point set is $\mathcal{T}(\mathbf{Y}) = s\mathbf{Y}\mathbf{R}^T + \mathbf{1}\mathbf{t}^T$,
- The probability of correspondence is given by $\mathbf{P}$.

Fig. 2. Rigid point set registration algorithm.

**Affine point set registration algorithm:**
- Initialization: $\mathbf{B} = \mathbf{I}, \mathbf{t} = 0, 0 \leq w \leq 1$
  $\sigma^2 = \frac{1}{DNM} \sum_{n=1}^{N} \sum_{m=1}^{M} \|\mathbf{x}_n - \mathbf{y}_m\|^2$
- EM optimization, repeat until convergence:
  - E-step: Compute $\mathbf{P}$,
  $$p_{mn} = \frac{\exp^{-\frac{1}{2\sigma^2}\|\mathbf{x}_n - (\mathbf{B}\mathbf{y}_m + \mathbf{t})\|^2}}{\sum_{k=1}^{M} \exp^{-\frac{1}{2\sigma^2}\|\mathbf{x}_n - (\mathbf{B}\mathbf{y}_k + \mathbf{t})\|^2} + (2\pi\sigma^2)^{D/2} \frac{w}{1-w} \frac{M}{N}}$$
  - M-step: Solve for $\mathbf{B}, \mathbf{t}, \sigma^2$:
    - $N_\mathbf{P} = \mathbf{1}^T \mathbf{P} \mathbf{1}, \mu_\mathbf{x} = \frac{1}{N_\mathbf{P}} \mathbf{X}^T \mathbf{P} \mathbf{1}, \mu_\mathbf{y} = \frac{1}{N_\mathbf{P}} \mathbf{Y}^T \mathbf{P} \mathbf{1}$,
    - $\hat{\mathbf{X}} = \mathbf{X} - \mathbf{1}\mu_\mathbf{x}^T, \hat{\mathbf{Y}} = \mathbf{Y} - \mathbf{1}\mu_\mathbf{y}^T$,
    - $\mathbf{B} = (\hat{\mathbf{X}}^T \mathbf{P}^T \hat{\mathbf{Y}})(\hat{\mathbf{Y}}^T d(\mathbf{P}\mathbf{1})\hat{\mathbf{Y}})^{-1}$,
    - $\mathbf{t} = \mu_\mathbf{x} - \mathbf{B}\mu_\mathbf{y}$,
    - $\sigma^2 = \frac{1}{N_\mathbf{P} D} (\mathrm{tr}(\hat{\mathbf{X}}^T d(\mathbf{P}^T \mathbf{1})\hat{\mathbf{X}}) - \mathrm{tr}(\hat{\mathbf{X}}^T \mathbf{P}^T \hat{\mathbf{Y}}\mathbf{B}^T))$.
- The aligned point set is $\mathcal{T}(\mathbf{Y}) = \mathbf{Y}\mathbf{B}^T + \mathbf{1}\mathbf{t}^T$,
- The probability of correspondence is given by $\mathbf{P}$.

Fig. 3. Affine point set registration algorithm.

Now we are ready to use Lemma 1, and the optimal $\mathbf{R}$ is in the form

$$\mathbf{R} = \mathbf{U}\mathbf{C}\mathbf{V}^T, \text{where } \mathbf{U}SS\mathbf{V}^T = svd(\hat{\mathbf{X}}^T \mathbf{P}^T \hat{\mathbf{Y}}) \quad (9)$$

and $\mathbf{C} = d(1,..,1, \det(\mathbf{U}\mathbf{V}^T))$. To solve for $s$ and $\sigma^2$, we equate the corresponding partial derivative of (8) to zero. Solving for $\mathbf{R}, s, \mathbf{t}, \sigma^2$ is the M-step of the EM algorithm. We summarize the rigid point sets registration algorithm in Fig. 2.

The algorithm has one free paramater, $w$ ($0 \leq w \leq 1$), which reflects our assumption on the amount of noise in the point sets. The solution for the rotation matrix is general $D$-dimensional.

**Affine point set registration:** Affine registration case is simpler compared to the rigid case, because the optimization is unconstrained. Affine transformation is defined as $\mathcal{T}(\mathbf{y}_m; \mathbf{R}, \mathbf{t}, s) = \mathbf{B}\mathbf{y}_m + \mathbf{t}$, where $\mathbf{B}_{D \times D}$ is an affine transformation matrix, $\mathbf{t}_{D \times 1}$ is translation vector. The objective function takes the form:

$$Q(\mathbf{B}, \mathbf{t}, \sigma^2) = \frac{1}{2\sigma^2} \sum_{m,n=1}^{M,N} P^{\mathrm{old}}(m|\mathbf{x}_n) \|\mathbf{x}_n - (\mathbf{B}\mathbf{y}_m + \mathbf{t})\|^2 + \frac{N_\mathbf{P} D}{2} \log \sigma^2 \quad (10)$$

We can directly take the partial derivatives of $Q$, equate them to zero, and solve the resulting linear system of equations. The solution is straightforward and similar to the rigid case. We summarize the affine point set registration algorithm in Fig. 3.

### 4.1 Related Rigid Point set Registration Methods

Here, we discuss the probabilistic rigid point set registration methods most closely related to ours. Rangarajan et al. [5] presented the RPM method for rigid point set registration. The method is shown for 2D case, where rotation matrix is parametrized by a single rotation angle, which allows to find an explicit update. Such Euler's angles approach is not feasible in multidimensional cases. RPM also uses deterministic annealing on $\sigma^2$, which requires to set the starting and stopping criteria as well as the annealing rate. The EM iterations has to be repeated at each annealing step, which can be slow. We argue that it is preferable to estimate $\sigma^2$ instead of using deterministic annealing. The deterministic annealing helps to overcome poor local minima, but for the rigid point set registration problem the rigid parametrization is a strong constraint that alleviates the advantages of the deterministic annealing.

Luo and Hancock [14], [36] introduced the rigid point sets registration algorithm that is the most similar to ours. The authors optimize the objective function rather intuitively than rigorously, which leads to several assumptions and approximate minimization. They ignore a few terms of the objective function (see Eqs.10,11 in [36]), where the last term does depend on transformation parameters, and must not be ignored. If such optimization converge, the M-step of the EM algorithm is only approximate. Among other differences, we want to mention that the authors use *structural editing*, a technique to remove some undesirable points, instead of using an additional uniform distribution to account for these points. Some other authors [15] also follow the rigid parameters estimation steps of Luo and Hancock [36].

## 5 NON-RIGID POINT SET REGISTRATION

Non-rigid point set registration remains a challenging problem in computer vision. The transformation that aligns the point sets is assumed to be unknown and non-rigid, which is generally broad class of transformations that can lead to an ill-posed problem. In order to deal with the problem we use Tikhonov regularization framework [37]–[39]. We define the transformation as the

initial position plus a displacement function $v$:

$$\mathcal{T}(\mathbf{Y}, v) = \mathbf{Y} + v(\mathbf{Y}), \quad (11)$$

We regularize the norm of $v$ to enforce the smoothness of the function [38]. Such approach is also supported by the Motion Coherence Theory (MCT) [29], [30], which states that points close to one another tend to move coherently, and thus, the displacement function between the point sets should be smooth. This is mathematically formulated as a regularization on the displacement (also called velocity) function.

Adding a regularization term to the negative log-likelihood function we obtain

$$f(v, \sigma^2) = E(v, \sigma^2) + \frac{\lambda}{2}\phi(v) \quad (12)$$

where $E$ is the negative log-likelihood function (3), $\phi(v)$ is a regularization term and $\lambda$ is a trade-off parameter. Such regularization is well formulated in Bayesian approach, where the regularization term comes from a prior on displacement field: $p(v) = \exp^{-\frac{\lambda}{2}\phi(v)}$.

We estimate the displacement function $v$ using variational calculus. We shall define the regularization term $\phi(v)$ in different but equivalent forms and show that the optimal functional form of $v$ is a linear combination of particular kernel functions. A particular choice of the regularization will lead to our non-rigid point set registration method.

### 5.1 Regularization of the Displacement Function

A norm of $v$ in the Hilbert space $\mathbb{H}^m$ is defined as:

$$\|v\|^2_{\mathbb{H}^m} = \int_{\mathbb{R}} \sum_{k=0}^{m} \left\| \frac{\partial^k v}{\partial x^k} \right\|^2 dx. \quad (13)$$

Alternatively, we can define the norm in the Reproducing Kernel Hilbert Space (RKHS) [38], [40] as

$$\|v\|^2_{\mathbb{H}^m} = \int_{\mathbb{R}^D} \frac{|\tilde{v}(\mathbf{s})|^2}{\tilde{G}(\mathbf{s})} d\mathbf{s} \quad (14)$$

where $G$ is a unique kernel function associated with the RKHS, and $\tilde{G}$ is its Fourier transform. Function $\tilde{v}$ indicates the Fourier transform of the function $v$ and $\mathbf{s}$ is a frequency domain variable. The Fourier domain norm definition has been used in the Regularization Theory (RT) [40] to regularize the smoothness of a function. Regularization theory defines smoothness as a measure of the "oscillatory" behavior of a function. Within the class of differentiable functions, one function is said to be smoother than another if it oscillates less; in other words, if it has less energy at high frequency. The high frequency content of a function can be measured by first high-pass filtering the function, and then measuring the resulting power. This can be represented by (14), where $\tilde{G}$ represents a symmetric positive definite low-pass filter, which approaches zero as $\|\mathbf{s}\| \to \infty$. For convenience, we shall write the regularization term as

$$\phi(v) = \|v\|^2_{\mathbb{H}^m} = \|Pv\|^2 \quad (15)$$

where an operator $P$ "extracts" a part of the function for regularization, in our case, the high frequency content part [38], [39].

### 5.2 Variational Solution

We find the functional form of $v$ using calculus of variation. Minimization of regularized negative log-likelihood function in (12) boils down to minimization of the following objective function (M-step):

$$Q(v, \sigma^2) = \frac{1}{2\sigma^2} \sum_{m,n=1}^{M,N} P^{\text{old}}(m|\mathbf{x}_n) \|\mathbf{x}_n - (\mathbf{y}_m + v(\mathbf{y}_m))\|^2$$
$$+ \frac{N_\mathbf{P} D}{2} \log \sigma^2 + \frac{\lambda}{2} \|Pv\|^2 \quad (16)$$

A function $v$ that minimizes (16) must satisfy the Euler-Lagrange differential equation

$$\frac{1}{\sigma^2 \lambda} \sum_{n=1}^{N} \sum_{m=1}^{M} P^{\text{old}}(m|\mathbf{x}_n)(\mathbf{x}_n - (\mathbf{y}_m + v(\mathbf{y}_m)))\delta(\mathbf{z} - \mathbf{y}_m)$$
$$= \hat{P} P v(\mathbf{z}) \quad (17)$$

for all vectors $\mathbf{z}$, where $\hat{P}$ is the adjoint operator to $P$. The solution to such partial differential equation can be written as the integral transformation of its left side with a Green's function $G$ of the self-adjoint operator $\hat{P}P$.

$$v(\mathbf{z}) = \frac{1}{\sigma^2 \lambda} \sum_{m,n=1}^{M,N} P^{\text{old}}(m|\mathbf{x}_n)(\mathbf{x}_n - (\mathbf{y}_m + v(\mathbf{y}_m)))G(\mathbf{z}, \mathbf{y}_m)$$
$$= \sum_{m=1}^{M} \mathbf{w}_m G(\mathbf{z}, \mathbf{y}_m) \quad (18)$$

where $\mathbf{w}_m = \frac{1}{\sigma^2 \lambda} \sum_{n=1}^{N} P^{\text{old}}(m|\mathbf{x}_n)(\mathbf{x}_n - (\mathbf{y}_m + v(\mathbf{y}_m)))$. Note that this solution is incomplete. In general, the solution also includes the term $\psi(\mathbf{z})$ that lies in the null space of $P$ [40], [41]. Thus, we achieve Lemma 2.

*Lemma 2:* The optimal displacement function that minimizes (16) is given by linear combination of the particular kernel functions centered at the points $\mathbf{Y}$ plus the term $\psi(\mathbf{z})$ in the null space of $P$:

$$v(\mathbf{z}) = \sum_{m=1}^{M} \mathbf{w}_m G(\mathbf{z}, \mathbf{y}_m) + \psi(\mathbf{z}) \quad (19)$$

where the kernel function is a Green's function of the self-adjoint operator $\hat{P}P$.

### 5.3 The Coherent Point Drift (CPD) Algorithm

We choose the regularization term according to (14):

$$\phi(v) = \int_{\mathbb{R}^D} \frac{|\tilde{v}(\mathbf{s})|^2}{\tilde{G}(\mathbf{s})} d\mathbf{s} \quad (20)$$

where $G$ is a Gaussian (note it is not related to the Gaussian form of the distribution chosen for the mixture model). There are several motivations for such a Gaussian choice: First, the Green's function (the kernel) corresponding to the regularization term in (20) is also a Gaussian (and remains a Gaussian for an arbitrary dimensional case); the Gaussian kernel is positive definite and the null space term $\psi(\mathbf{z}) = 0$ [40]. Second, by choosing an appropriately sized Gaussian function we have the flexibility to control the range of filtered frequencies and thus the amount of spatial smoothness. Third, the choice of the Gaussian makes our regularization term equivalent to the one in the Motion Coherence Theory (MCT) [30]:

$$\phi_{MCT}(v) = \int_{\mathbb{R}^d} \sum_{l=0}^{\infty} \frac{\beta^{2l}}{l!2^l} \left\| D^l v(\mathbf{x}) \right\|^2 d\mathbf{x} \quad (21)$$

where $D$ is a derivative operator so that $D^{2l}v = \nabla^{2l}v$ and $D^{2l+1}v = \nabla(\nabla^{2l}v)$, where $\nabla$ is the gradient operator and $\nabla^2$ is the Laplacian operator.

*Lemma 3:* The regularization term in (20) with a Gaussian choice of low-pass filter $G$ is equivalent to the the regularization term in (21). Both terms represent the norm of the function $v$, after applying the operator $P$, and the corresponding Green's function is a Gaussian in both cases [38].

The equivalence of our regularization term with that of the Motion Coherence Theory implies that we are imposing motion coherence among the points and thus we call the non-rigid point set registration method the Coherent Point Drift (CPD) algorithm.

We can obtain the coefficients $\mathbf{w}_m$ by evaluating (19) at $\mathbf{y}_m$ points

$$(\mathbf{G} + \lambda\sigma^2 d(\mathbf{P1})^{-1})\mathbf{W} = d(\mathbf{P1})^{-1}\mathbf{PX} - \mathbf{Y} \quad (22)$$

where $\mathbf{W}_{M \times D} = (\mathbf{w}_1, \ldots, \mathbf{w}_M)^T$ is a matrix of coefficients, $\mathbf{G}_{M \times M}$ is a kernel matrix with elements $g_{ij} = G(\mathbf{y}_i, \mathbf{y}_j) = e^{-\frac{1}{2}\left\|\frac{\mathbf{y}_i - \mathbf{y}_j}{\beta}\right\|^2}$ and $d^{-1}(\cdot)$ is the inverse diagonal matrix. The transformed position of $\mathbf{y}_m$ are found according to (11) as $\mathbf{T} = \mathcal{T}(\mathbf{Y}, \mathbf{W}) = \mathbf{Y} + \mathbf{GW}$. We obtain $\sigma^2$ by equating the corresponding derivative of $Q$ to zero

$$\sigma^2 = \frac{1}{N_\mathbf{P} D} \sum_{n=1}^{N} \sum_{m=1}^{M} \|\mathbf{x}_n - \mathcal{T}(\mathbf{y}_m, \mathbf{W})\|^2 =$$
$$\frac{1}{N_\mathbf{P} D}(\operatorname{tr}(\mathbf{X}^T d(\mathbf{P}^T \mathbf{1})\mathbf{X}) - 2\operatorname{tr}((\mathbf{PX})^T \mathbf{T}) + \operatorname{tr}(\mathbf{T}^T d(\mathbf{P1})\mathbf{T}))$$
(23)

We summarize the CPD non-rigid point set registration algorithm in Fig. 4.

**Analysis:** The CPD algorithm includes three free parameters: $w, \lambda$ and $\beta$. Parameter $w$ ($0 \leq w \leq 1$) reflects our assumption on the amount of noise in the point sets. Parameters $\lambda$ and $\beta$ both reflect the amount of smoothness regularization. A discussion on the difference between $\lambda$ and $\beta$ can be found in [29], [30]. Briefly speaking, parameter $\beta$ defines the model of the smoothness regularizer (width of smoothing Gaussian filter in (20)). Parameter $\lambda$ represents the trade-off between the goodness of maximum likelihood fit and regularization.

---

**Non-rigid point set registration algorithm:**
- Initialization: $\mathbf{W} = 0, \sigma^2 = \frac{1}{DNM} \sum_{m,n=1}^{M,N} \|\mathbf{x}_n - \mathbf{y}_m\|^2$
- Initialize $w (0 \leq w \leq 1)$, $\beta > 0$, $\lambda > 0$,
- Construct $\mathbf{G}$: $g_{ij} = \exp^{-\frac{1}{2\beta^2}\|\mathbf{y}_i - \mathbf{y}_j\|^2}$,
- EM optimization, repeat until convergence:
  - E-step: Compute $\mathbf{P}$,
  $$p_{mn} = \frac{\exp^{-\frac{1}{2\sigma^2}\|\mathbf{x}_n - (\mathbf{y}_m + \mathbf{G}(m,\cdot)\mathbf{W})\|^2}}{\sum_{k=1}^{M} \exp^{-\frac{1}{2\sigma^2}\|\mathbf{x}_n - (\mathbf{y}_k + \mathbf{G}(k,\cdot)\mathbf{W})\|^2} + \frac{w}{1-w}\frac{(2\pi\sigma^2)^{D/2}M}{N}}$$
  - M-step:
    · Solve $(\mathbf{G} + \lambda\sigma^2 d(\mathbf{P1})^{-1})\mathbf{W} = d(\mathbf{P1})^{-1}\mathbf{PX} - \mathbf{Y}$
    · $N_\mathbf{P} = \mathbf{1}^T\mathbf{P1}$, $\mathbf{T} = \mathbf{Y} + \mathbf{GW}$,
    · $\sigma^2 = \frac{1}{N_\mathbf{P} D}(\operatorname{tr}(\mathbf{X}^T d(\mathbf{P}^T\mathbf{1})\mathbf{X}) - 2\operatorname{tr}((\mathbf{PX})^T\mathbf{T}) + \operatorname{tr}(\mathbf{T}^T d(\mathbf{P1})\mathbf{T}))$,
- The aligned point set is $\mathbf{T} = \mathcal{T}(\mathbf{Y}, \mathbf{W}) = \mathbf{Y} + \mathbf{GW}$,
- The probability of correspondence is given by $\mathbf{P}$.

Fig. 4. The Coherent Point Drift algorithm for non-rigid point set registration.

We note that solution of (22) gives the exact minimum of $Q$ (16), if $\sigma^2$ is assumed fixed. As far as we are estimating $\sigma^2$, (22) and (23) should be solved simultaneously. The non-linear dependency of $\sigma^2$ on $\mathbf{W}$ and vice-verse does not allow for simultaneous analytical solution. Iterative exact solution can be obtained by performing a few cyclic iterations on (22) and (23) within a single EM step. Practically, a single iteration, given by (22) and (23), decrease the $Q$ function almost to the exact minimum. Such an iterative procedure, which decreases the $Q$ function but not to exact minimum, is often called the generalized EM algorithm [31], [42].

### 5.4 Related Non-rigid Point set Registration Methods

The CPD algorithm follows our previous work [28] on non-rigid point sets registration. However, previously we have used deterministic annealing on $\sigma^2$, whereas here, we estimate the Gaussian width $\sigma^2$ within ML framework. This allows us to significantly speed up the algorithm, alleviating the repeated EM-iterations for every single annealing step. We have not observed any decrease in accuracy of the method related to this change. In [28], we used a slightly different notation for the GMM centroid locations: we called $\mathbf{Y}_0$ the initial centroids position (which we call $\mathbf{Y}$ here), and $\mathbf{Y}$ for the finall GMM centroid position (which we call $\mathcal{T}(\mathbf{Y})$ here).

The most relevant non-rigid point sets registration algorithm to ours is TPS-RPM, more precisely its GMM formulation [10]. TPS-RPM uses Thin Plate Spline (TPS) [23], [24] parametrization of the transformation, which can be obtained by adding the regularization term that penalizes second order derivatives of the transfor-mation. For instance, in 2D such regularization term is

$$\|L\mathcal{T}\|^2 = \int\int[(\frac{\partial^2\mathcal{T}}{\partial x^2})^2 + 2(\frac{\partial^2\mathcal{T}}{\partial x\partial y})^2 + (\frac{\partial^2\mathcal{T}}{\partial y^2})^2]dxdy \quad (24)$$

This term can be equivalently formulated in the Fourier space as:

$$\|L\mathcal{T}\|^2 = \int_{\mathbb{R}^2} \|\mathbf{s}\|^4 |\tilde{\mathcal{T}}(\mathbf{s})|^2 d\mathbf{s} \quad (25)$$

which is a special case of the Duchon splines [43]. The null space of such regularization includes affine transformations. Using the variational approach we can show that the optimal transformation $\mathcal{T}$ for such regularization is in the form $\mathcal{T}(\mathbf{Y}) = \mathbf{YA} + \mathbf{KC}$, where $\mathbf{A}$ is matrix of affine transformation cooefficients, $\mathbf{C}$ is a matrix of non-rigid cooefficients. For 2D case, matrix $\mathbf{K}_{M\times M}$ is kernel matrix with elements $k_{ij} = \|\mathbf{y}_i - \mathbf{y}_j\|^2 \log \|\mathbf{y}_i - \mathbf{y}_j\|$. For 3D case, matrix $\mathbf{K}$ has elements $k_{ij} = \|\mathbf{y}_i - \mathbf{y}_j\|$. For 4D or higher dimensions the TPS kernel solution does not exist [44]. Finally, to link such regularization to our non-rigid registration framework, we note that the regularization of the displacement field $v$, instead of the transformation itself, is exactly the same, because, (24) is invariant under affine transformations, in other words $\|L\mathcal{T}(\mathbf{z})\|^2 = \|L(\mathbf{z}+v(\mathbf{z}))\|^2 = \|Lv(\mathbf{z})\|^2$. This means that both CPD and TPS-RPM regularizes the displacement function, but using different regularization terms.

The advantage of CPD regularization (as given by (20) or (21)) comparing to TPS ((24) or (25)), is that it easily generalizes to $N$ dimensions. Also we can control the locality of spatial smoothness by changing the Gaussian filter width $\beta$, whereas TPS does not have such flexibility. This, however, introduces one extra parameter to the method, but TPS-RPM has to uses one extra parameter to regularize affine matrix after all. Among other differences, TPS-RPM approximates the M-step solution of the EM algorithm [10] for simplicity and use deterministic annealing on $\sigma^2$.

Finally, Jian and Vemuri [26] consider the registration as an alignment between the distributions of two point sets, where a separate GMMs are used to model the distribution for the point sets. One of the point sets is parametrized by TPS. The transformation parameters are estimated to minimize the $L_2$ norm between the distributions. In our case, the CPD method maximizes the likelihood function, which is equivalent to KL divergence minimization between two mixture distributions: GMM and mixture of delta functions. KL divergence is more appropriate similarity measure for the densities than $L_2$ norm, because it weights the error according to its probability.

## 6 Fast Implementation

Here we show that CPD computational complexity can be reduced to linear up to a multiplicative constant. We use the fast Gauss transform (FGT) [45] to compute the matrix-vector products $\mathbf{P1}, \mathbf{P}^T\mathbf{1}, \mathbf{PX}$, which are the bottlenecks for both rigid and non-rigid cases. We use

> **Compute $\mathbf{P}^T\mathbf{1}, \mathbf{P1}$ and $\mathbf{PX}$:**
> - Compute $\mathbf{K}^T\mathbf{1}$ (using FGT),
> - $\mathbf{a} = 1./(\mathbf{K}^T\mathbf{1} + c\mathbf{1})$,
> - $\mathbf{P}^T\mathbf{1} = \mathbf{1} - c\mathbf{a}$,
> - $\mathbf{P1} = \mathbf{Ka}$ (using FGT),
> - $\mathbf{PX} = \mathbf{K}(\mathbf{a}.*\mathbf{X})$ (using FGT),

Fig. 5. Matrix-vector products computation through FGT.

low-rank matrix approximation to speed-up the solution of the linear system of equations (22) for the non-rigid case.

**The fast Gauss transform:** Greengard and Strain [45] introduced the fast Gauss transform (FGT) for fast computation of the sum of exponentials:

$$f(\mathbf{y}_m) = \sum_{n=1}^{N} z_n \exp^{-\frac{1}{2\sigma^2}\|\mathbf{x}_n - \mathbf{y}_m\|^2}, \quad \forall \mathbf{y}_m, \ m = 1, \ldots, M.$$

The naive approach takes $\mathcal{O}(MN)$ operations, while FGT takes only $\mathcal{O}(M+N)$. The basic idea of FGT is to expand the Gaussians in terms of truncated Hermit expansion, and approximate (6) up to the predefined accuracy. Rewriting (6) in matrix form, we obtain $\mathbf{f} = \mathbf{Kz}$, where $\mathbf{z}$ is some vector and $\mathbf{K}_{M\times N}$ is a Gaussian affinity matrix with elements: $k_{mn} = \exp^{-\frac{1}{2\sigma^2}\|\mathbf{x}_n - \mathcal{T}(\mathbf{y}_m)\|^2}$, which we have already used in our notations. We simplify the matrix-vector products $\mathbf{P1}, \mathbf{P}^T\mathbf{1}$ and $\mathbf{PX}$, to the form of $\mathbf{Kz}$ and apply FGT. Matrix $\mathbf{P}$ (6) can be partitioned into

$$\mathbf{P} = \mathbf{K}\,\mathrm{d}(\mathbf{a}), \quad \mathbf{a} = 1./(\mathbf{K}^T\mathbf{1} + c\mathbf{1}) \quad (26)$$

where $\mathrm{d}(\mathbf{a})$ is diagonal matrix with a vector $\mathbf{a}$ along the diagonal. Here, we use Matlab element-wise division (./) and element-wise (.$*$) multiplication notations. We show the algorithm to compute the bottleneck matrix-vector products $\mathbf{P1}, \mathbf{P}^T\mathbf{1}$ and $\mathbf{PX}$ using FGT in Fig. 5. We note that for dimensions higher than three, we can use the improved fast Gauss transform (IFGT) method [46], which is a faster alternative to FGT for higher dimensions.

During the finall EM iterations, the width of the Gaussians $\sigma^2$ becomes small. The Hermitian expansion thus requires many terms to approximate highly multimodal Gaussian distribution for a given precision. At the final iterations, the Gaussian becomes very narrow, and we can switch to the truncated Gaussian approximation (set zeros outside a predefined box).

**Low-rank matrix approximation:** In the non-rigid case, we need to solve the linear system (22), which is $\mathcal{O}(M^3)$ using direct matrix inversion. We note that the left hand side matrix of (22) is symmetric and positive definite.

We use low-rank matrix approximation of $\mathbf{G}$, where $\mathbf{G}$ is a Gaussian affinity matrix with elements $g_{ij} = \exp^{-\frac{1}{2\beta^2}\|\mathbf{y}_i - \mathbf{y}_j\|^2}$. We approximate the matrix $\mathbf{G}$ as

$$\hat{\mathbf{G}} = \mathbf{Q}\Lambda\mathbf{Q}^T \quad (27)$$

where $\Lambda_{K\times K}$ is a diagonal matrix with $K$ largest eigenvalues and the matrix $\mathbf{Q}_{M\times K}$ is formed from the cor-responding eigenvectors. $\hat{\mathbf{G}}$ is the closest $K$-rank matrix approximation to $\mathbf{G}$ both in $L_2$ and Frobenius norms [47]. To solve the linear system in (22) we use the Woodbury identity and invert the first term as

$$(\mathbf{Q}\Lambda\mathbf{Q}^T + \lambda\sigma^2\,\mathrm{d}(\mathbf{P1})^{-1})^{-1} = \frac{1}{\lambda\sigma^2}\,\mathrm{d}(\mathbf{P1})$$
$$-\frac{1}{(\lambda\sigma^2)^2}\,\mathrm{d}(\mathbf{P1})\mathbf{Q}(\Lambda^{-1} + \frac{1}{\lambda\sigma^2}\mathbf{Q}^T\,\mathrm{d}(\mathbf{P1})\mathbf{Q})^{-1}\mathbf{Q}^T\,\mathrm{d}(\mathbf{P1})$$
(28)

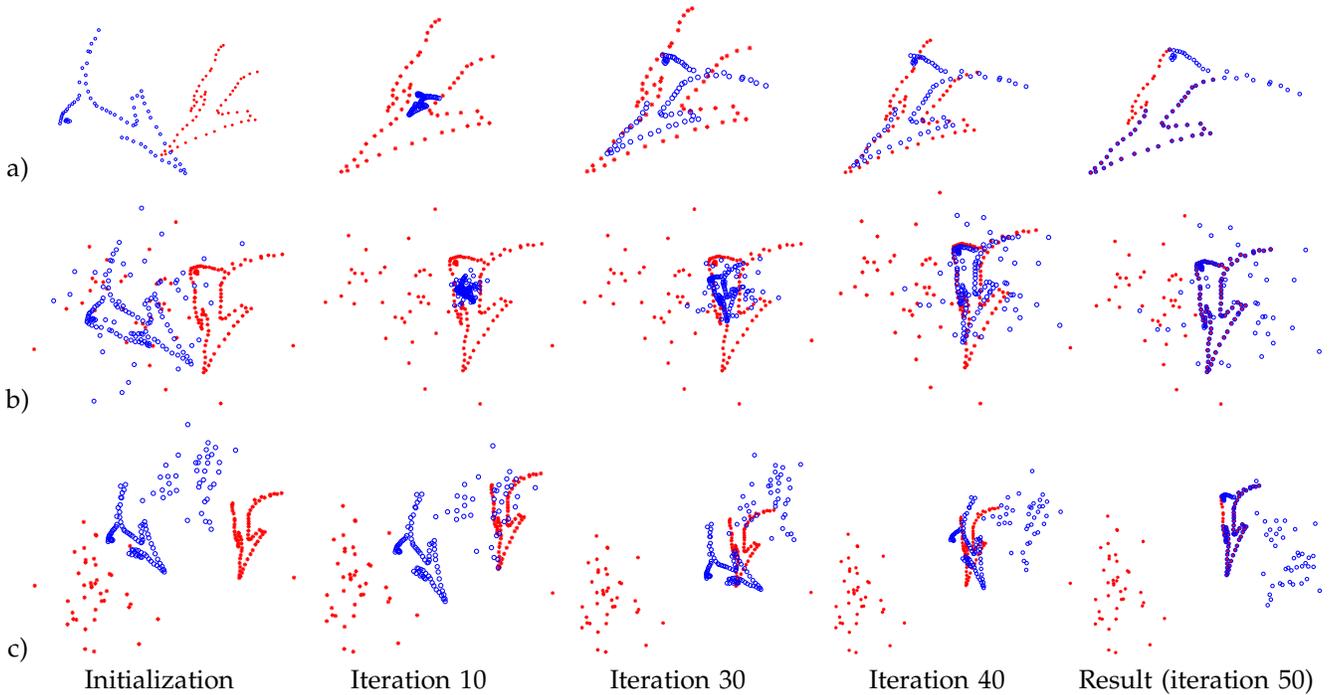

Fig. 6. Fish data set, rigid registration examples. We align $\mathbf{Y}$ (blue circles) onto $\mathbf{X}$ (red stars). The columns show the iterative alignment progress. a) Registration of the point sets with missing non-overlapping parts ($w = 0.5$); b) Registration of the point sets corrupted by random outliers ($w = 0.5$); c) A challenging rigid registration example, where both point sets are corrupted by outliers and biased to different sides of the point sets. We have also deleted some parts from both point sets. We set $w = 0.8$ and fix scaling $s = 1$. CPD registration is robust and accurate in all experiments.

The inside matrix inversion is of $\mathcal{O}(K^3)$, where $K \ll M$. For instance choosing $K = M^{1/3}$ largest eigenvalues, we reduce the computational complexity to linear. We can pre-compute $K$ largest eigenvalues and eigenvectors of $\mathbf{G}$ using deflation techniques [48]. It requires several iterations with the matrix-vector product $\mathbf{Gz}$, which can be implemented explicitly or through FGT.

The low-rank matrix approximation intuitively constraints the space of the non-rigid transformations, and can be even desirable to further constrain the non-rigid transformation. If the number of points is large and well clustered, then an extremely small percent of eigenvalues will be sufficient for an accurate approximation.

## 7 RESULTS

We implemented the algorithm in Matlab, and tested it on a Pentium4 CPU 3GHz with 4GB RAM. We implemented the matrix-vector products in C as a Matlab mex functions to avoid the storage of $\mathbf{P}$. The code is available at www.csee.ogi.edu/~myron/matlab/cpd. We shall refer to our method as Coherent Point Drift (CPD) both for rigid and non-rigid point sets registration methods presented in this paper. We have also implemented the matrix-vector products through FGT using the Matlab FGT implementation by Sebastien Paris [49].

We consider rigid and non-rigid experiments separately below. We shall always pre-align both point sets to zero mean and unit variance before the registration.

### 7.1 Rigid Registration Results

We show the CPD rigid registration on several examples, test the fast CPD implementation and evaluate its performance in comparison with LM-ICP [3], which is one of the most popular robust rigid point set registration methods.

**Rigid fish point set registration:** Fig. 6 shows several rigid regsitration tests on 2D fish point sets. In Fig. 6a we deleted non-overlapping parts in both point sets and set $w = 0.5$, where $w$ is a weight of the uniform ditribution that accounts for noise and outliers. In Fig. 6b we corrupted the point sets by outliers. We generate outliers randomly from a normal zero-mean distribution. CPD demonstrates robust and accurate performance in all examples. Fig. 6c demonstrates a challenging example, where both point sets have missing points and are corrupted by outliers. The most challenging here is that we biased the outliers to the different sides of fish point sets. We were able to register such point sets only by fixing the scaling to be constant (estimating rotation and translation only). CPD demonstrates accurate and robust registration performance.

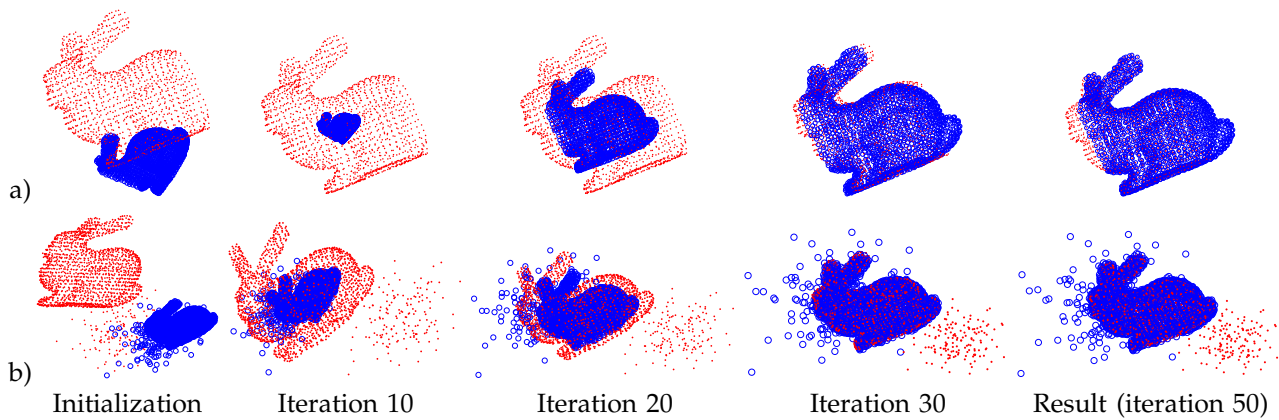

|  |  |  |  |  |
|---|---|---|---|---|
| Initialization | Iteration 10 | Iteration 20 | Iteration 30 | Result (iteration 50) |

Fig. 7. 3D bunny point set rigid registration examples. We align $\mathbf{Y}$ (blue circles) onto $\mathbf{X}$ (red dots). The columns show the iterative alignment progress. We initialized one of the point sets with 50 degree rotation and scaling equal $2$. a) Registration of the point sets with missing points ($w = 0.5$); b) A challenging example of CPD rigid registration with missing points, outliers and noise. CPD shows robust and accurate registration result in all experiments.

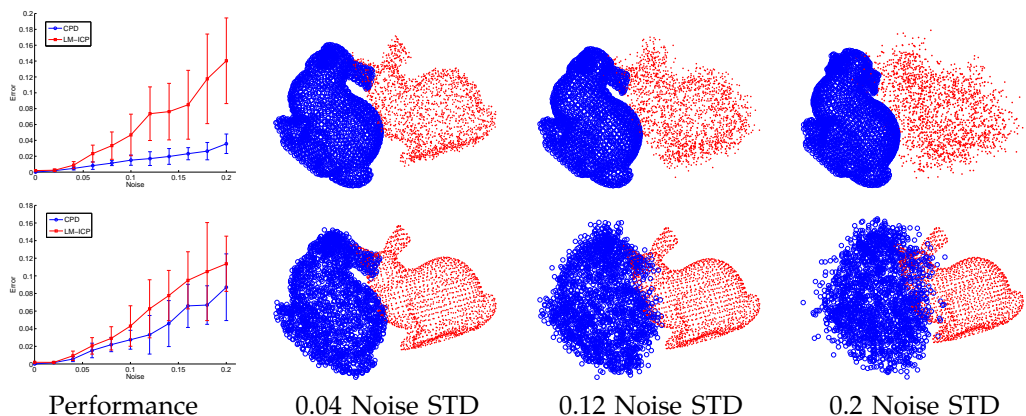

| Performance | 0.04 Noise STD | 0.12 Noise STD | 0.2 Noise STD |
|---|---|---|---|

Fig. 8. A comparison of CPD and LM-ICP rigid registration performances with respect to noise in the $\mathbf{X}$ (first row) and the $\mathbf{Y}$ point sets (second row). We align $\mathbf{Y}$ (blue circles) onto $\mathbf{X}$ (red dots). The columns 2,3 and 4 show the examples of initial point sets for different random noise stds added to the point set positions. The first column shows the error in estimating the rotation matrix for CPD (blue) and TPS-RPM (red). CPD outperforms LM-ICP in all cases.

**Rigid bunny point set registration:** We test 3D rigid point sets registration on the Stanford "bunny" data set [50]. We use a subsampled bunny version of $1889 \times 3$ points. In Fig. 7a, we have deleted the front and back parts of the bunny point sets. In Fig. 7b, we have added random outliers to different sides of the point sets. We set $w = 0.7$. CPD registration is accurate and robust in all examples.

We compare the CPD rigid algorithm to the LM-ICP method [3], a robust version of ICP. Fig. 8 shows the performance of CPD and LM-ICP with respect to noise in the point sets. We align the $\mathbf{Y}$ point set (blue circles) onto the $\mathbf{X}$ point set (red dots). We set $w = 0.5$. The known initial rotation discrepancy between the point sets is $50$ degrees. The first and second rows shows the alignment performance when a random noise is added to the $\mathbf{X}$ and $\mathbf{Y}$ point set positions respectively. We use a norm of the difference between the true and estimated rotation matrix as an error measure. A few initial point sets examples with different noise std are shown in the columns 2, 3 and 4 of Fig. 8. For each level of the noise stds we made $25$ independent runs. The first column plots the error values (mean and standard deviation) in the estimated rotation matrix as a function of noise levels. The CPD rigid algorithm outperforms the robust LM-ICP method, especially when the noise is present in the $\mathbf{X}$ point set.

Fig. 9 shows the performance of CPD and LM-ICP with respect to the outliers in the point sets. We add different number of outliers (irrelevant random points) to the point sets. An examples of such initial point sets are shown in columns 2, 3 and 4 of Fig. 9 for 600, 1800 and 3000 outlier points added respectively. The first and second row show the cases of outliers present in the $\mathbf{X}$ and $\mathbf{Y}$ point sets respectively. CPD performs well in all experiments, whereas LM-ICP performance is less accurate.

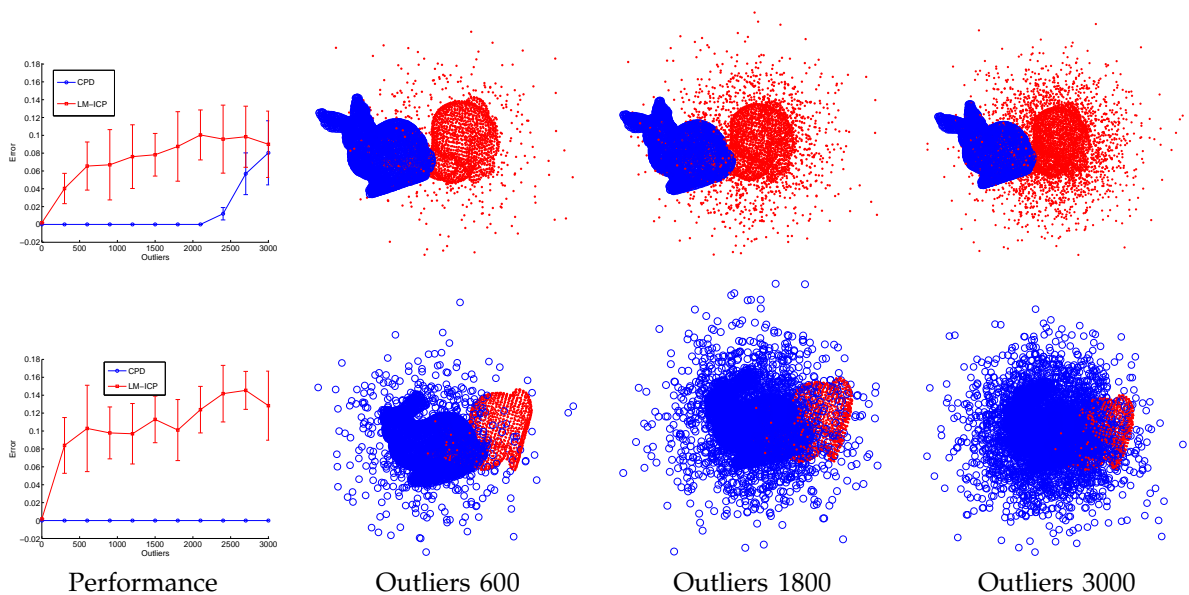

Fig. 9. A comparison of CPD and LM-ICP rigid registration performances with respect to outliers in the $\mathbf{X}$ (first row) and the $\mathbf{Y}$ (second row) point sets. We align $\mathbf{Y}$ (blue circles) onto $\mathbf{X}$ (red dots). The columns 2,3 and 4 show the examples of initial point sets with different number of outliers added. The first column show the error in estimating the rotation matrix. CPD outperforms LM-ICP.

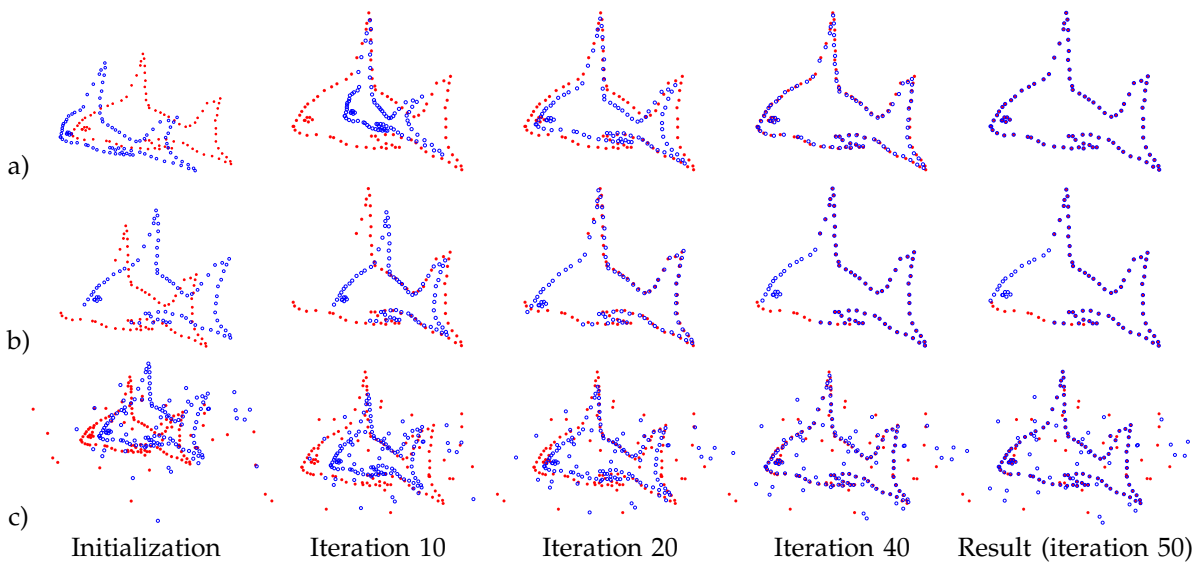

Fig. 10. Non-rigid CPD registration of 2D fish point sets. a) Noiseless fish point sets registration ($91 \times 2$ points, $w = 0$); b) Registration of 2D fish point set with missing points ($w = 0.5$); c) Registration of 2D fish point set in presence of outliers ($w = 0.5$). CPD registration is robust and accurate in all experiments.

**Fast rigid CPD implementation:** We also test the CPD performance with FGT implementation of the matrix-vector products. We use four Stanford bunny sets of sizes: $453 \times 3$, $1889 \times 3$, $8171 \times 3$ and $35947 \times 3$. For each of the cases we add a small amount of noise and outliers to both point sets, initialized them with 50 degrees rotation and set $w = 0.3$. For the FGT parameters, we used "ratio of far field"=8, "number of centers"=80, "order of truncation"=5. Table. 1 shows the registration time with and without FGT. The FGT implementation is significantly faster. We note that there are several downsides of using the FGT: a) FGT requires its own parameter initialization; b) CPD (with FGT) aligns the point sets to 0.1 degree error rotation and then starts being unstable. This is because $\sigma^2$ becomes small and the FGT approximation error becomes significant. At this point one can either stop (the alignment already is reasonably accurate) or proceed with ICP or truncated Gaussian CPD.

12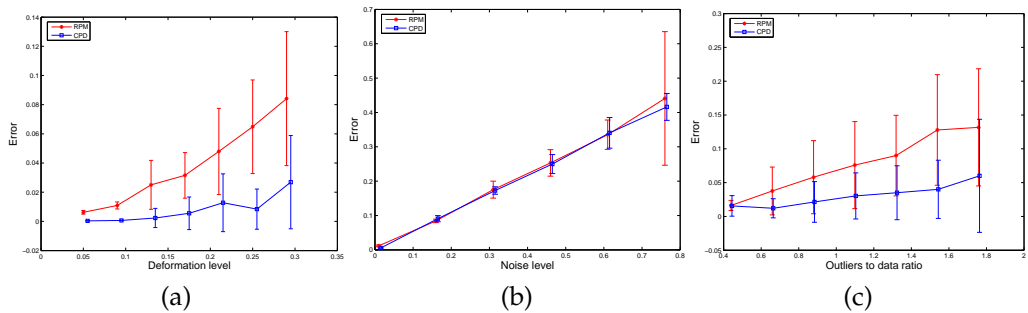

Fig. 11. A comparison of CPD and TPS-RPM on the 2D fish point sets with respect to a) Deformation level; b) Noise level; c) Outliers. CPD shows more accurate registration performance compared to TPS-RPM, especially in presence of outliers and complex non-rigid deformations.

| $N, M$ | Naive | FGT |
|---|---|---|
| $453 \times 3$ | 0.6s | 0.7s |
| $1889 \times 3$ | 11s | 3s |
| $8171 \times 3$ | 4m | 10s |
| $35947 \times 3$ | 3.5hr | 51s |

TABLE 1
The rigid CPD registration time for naive (no FGT) and FGT implementations. The FGT-based implementation is significantly faster.

| $N, M \times D$ | Naive | FGT | Low-rank | FGT & Low-rank |
|---|---|---|---|---|
| $453 \times 3$ | 2s | 2.3s | 1.7s | 1.8s |
| $1889 \times 3$ | 1m22s | 1m16s | 19s | 11s |
| $8171 \times 3$ | 3hr | 2hr26m | 10m20s | 1m37s |
| $35947 \times 3$ | – | – | 40m | 10m |

TABLE 2
Registration time required for non-rigid registration of 3D bunny point sets. The time is shown when using only FGT of vector-matrix products, only low-rank matrix approximation of Gaussian kernel matrix or both.

## 7.2 Non-rigid Registration Results

We show CPD non-rigid registration on several examples, test the fast CPD implementation and evaluate CPD performance in comparison to TPS-RPM [9], which is one of the best performing non-rigid point set registration methods. We set $\lambda = 2$, $\beta = 2$.

**Non-rigid fish point set registration:** Fig. 10a shows non-rigid CPD registration of two fish point sets with clean data. Fig. 10b is with missing points ($w = 0.5$). Fig. 10c is with both point sets are corrupted by outliers ($w = 0.5$). The non-rigid CPD registration results are accurate in all experiments.

We test CPD against TPS-RPM [9] on synthetically generated 2D fish non-rigid examples with respect to a) level of non-rigid deformation, b) amount of noise in the point sets locations c) number of outliers. We set $w = 0.3$ in all experiments. Since we know the true correspondences, we use the mean squared distance between the corresponding points after the registration as an error measure. For each set of parameters we have conducted 25 runs. Fig. 11a shows the methods performances with respect to the level of initial non-rigid deformation between the point sets. To generate the non-rigid transformation, we parameterize the point sets domain by a mesh of control points, perturb the points and use splines to interpolate the deformation. The higher level of mesh point perturbations produce the higher deformation. CPD shows accurate registration performance and outperforms the TPS-RPM. Fig. 11b shows the methods performances with respect to the amount of noise. We add a zero-mean white noise with increasing levels of stds to the point sets. Both CPD and TPS-RPM show accurate performances. We note that, due to deterministic annealing used by TPS-RPM, its convergence takes significantly more iterations and time. Fig. 11a shows the methods performances with respect to the number of outliers. We add random outliers to the point sets and plot the registration error with respect to the ration of number of outliers to the number of data points. CPD shows robust registration performance and outperforms the TPS-RPM.

**Non-rigid 3D face registration:** We show the CPD performance on 3D face point sets. Fig. 12a shows two 3D face point sets related through non-rigid deformation. Fig. 12b shows two 3D face point sets point sets with added outliers and non-rigid deformation. Non-rigid CPD registration is accurate in all experiments.

**Non-rigid 3D LV point set registration:** Finally, we demonstrate the CPD performance on non-rigid a 3D left ventricle (LV) contours segmented from 3D ultrasound images, using active contour based segmentation [51]. Fig. 13 shows (a) two LV point sets at different time instances, (b) the registration result, (d) the displacement field required for CPD alignment. That the registration result is accurate.

**Fast non-rigid CPD implementation:** We test the computational time of the fast CPD non-rigid implementation on several subsampled 3D Stanford bunny point sets. We use FGT of the matrix-vector products, the low-rank matrix approximations of the kernel matrix, or both. We applied a moderate non-rigid deformation to the bunny point sets. The registration time of the non-rigid CPD is shown in Table 2. We were unable to run the test without the low-rank matrix approximation for the largest bunny set ($35947 \times 3$), because of the large RAM requirements to construct the kernel matrix $\mathbf{G}$. We used only $100$ leading eigenvalues and eigenvectors in all cases. Table 2 shows that the main computational bottleneck is in solving the linear system of equations (22), because the low-rank matrix approximation alone can reduce the computational time significantly. Both FGT and low-rank approximations provide further speed-up with only moderate loss of accuracy. We note that almost $60\%$ of the time required to complete the CPD registration using the low-rank matrix approximation were required to pre-compute the eigenvalues and eigenvectors of the kernel matrix $\mathbf{G}$.

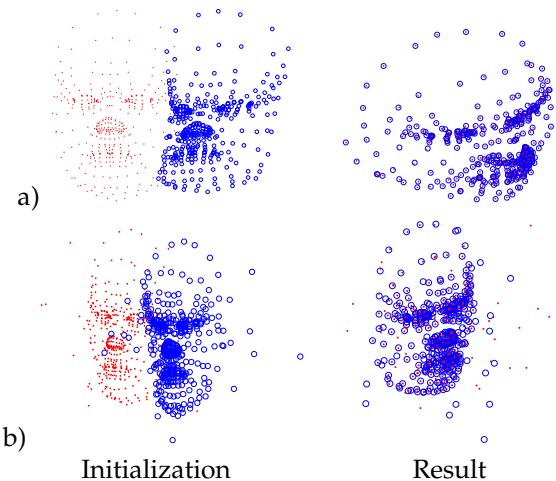

Fig. 12. Non-rigid registration of 3D face point sets. a) Registration of clean point sets b) Registration of point set with outliers. CPD shows accurate alignment.

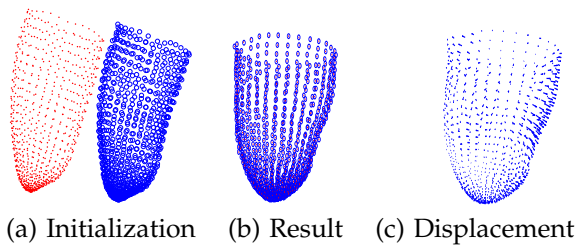

(a) Initialization  (b) Result  (c) Displacement

Fig. 13. Non-rigid registration of 3D left ventricle (LV) point sets. (a) two LV point sets at different time instances, (b) the registration result, (c) displacement field between the corresponding points.

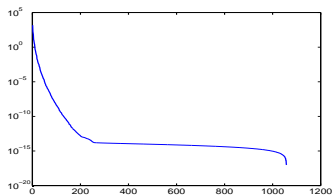

Fig. 14. The log-plot of the eigenspectrum of the kernel matrix $\mathbf{G}$ for the bunny point sets of size $1889 \times 3$.

We also show the eigenvalues for a particular example of the bunny point set of size $1889 \times 3$ in Fig. 14. Eigenvalues drops quickly below $10^{-3}$ only after 10 largest eigenvalues, and drops below $10^{-5}$ after about 100 eigenvalues. The approximation error of using a low rank approximate matrix (constructed from $100$ leading eigenvectors and eigenvalues), is only $10^{-8}$ in terms of its Frobenius norm.

## 8 Discussion and Conclusion

We introduce a probabilistic method for rigid and non-rigid point set registration, called the Coherent Point Drift algorithm. We consider the alignment of two point sets as a probability density estimation, where one point set represents the Gaussian Mixture Model centroids, and the other represents the data points. We iteratively fit the GMM centroids by maximizing the likelihood and find the posterior probabilities of centroids, which provide the correspondence probability. Core to our method is to force the GMM centroids to move coherently as a group, which preserves the topological structure of the point sets.

Our contribution includes the following aspects. For the non-rigid point set registration, we formulate the motion coherence constraint and derive a solution of the regularized ML estimation through the variational approach, which leads to an elegant kernel form. CPD simultaneously finds both the transformation and the correspondence between two point sets without making any prior assumption of the non-rigid transformation model except that of motion coherence. For the rigid case, we derived a closed form multidimensional solution (of the M-step of the EM algorithm), which has not been derived exactly before. Finally, we introduced the fast CPD implementation using fast Gauss transform and low-rank matrix approximation to reduce the computational complexity of the method to as low as linear. On top of the computational advantage, the low-rank kernel approximation provides more stable solutions in cases where the matrix $\mathbf{G}$ is poorly conditioned. To our best knowledge, CPD is the only method capable of non-rigid registration of large data sets. Both rigid and non-rigid CPD registration methods are multidimensional and can be applied to arbitrary dimensional data sets.

We estimate the GMM width, $\sigma^2$, within the ML formulation. We have not observed any decrease in performance compared to the deterministic annealing approach. Estimation $\sigma^2$ allows to reduce the number of free parameters and, most importantly, to significantly reduce the number of iterations and the processing time.

We have used an addition uniform distribution to account for noise and outliers. The weight, $w$, of this distribution provides a flexible control over the method robustness and allows accurate CPD performance, especially in presence of severe outliers and missing points.

We have tested CPD on various synthetic and real examples and comare it to LM-ICP (in rigid case) and TPS-RPM (in non-rigid case). CPD shows robust and accurate performance with respect to noise, outliers and missing points. Our method is of general interest with numerous computer vision applications. We provide the Matlab code of the CPD algorithm free for academic research.